\documentclass[letterpaper, 10 pt, conference]{ieeeconf}  

\IEEEoverridecommandlockouts                              

\usepackage{cite}
\usepackage{amsmath,amssymb,amsfonts}
\usepackage{algorithmic}
\usepackage{graphicx}
\usepackage{textcomp}
\usepackage{xcolor}
\usepackage{bbold}
\usepackage[numbers]{natbib}

\usepackage{hyperref}

\title{\LARGE \bf
Improvement of Optimization using Learning Based Models in \\Mixed Integer Linear Programming Tasks\vspace{-4mm}}

\author{Xiaoke Wang*, Batuhan Altundas*, Zhaoxin Li*, Aaron Zhao, and Matthew Gombolay
\thanks{*These authors have contributed equally.}
}

\begin{document}

\maketitle
\thispagestyle{empty}
\pagestyle{empty}

\begin{abstract}

Mixed Integer Linear Programs (MILPs) are essential tools for solving planning and scheduling problems across critical industries such as construction, manufacturing, and logistics. However, their widespread adoption is limited by long computational times, especially in large-scale, real-time scenarios. To address this, we present a learning-based framework that leverages Behavior Cloning (BC) and Reinforcement Learning (RL) to train Graph Neural Networks (GNNs), producing high-quality initial solutions for warm-starting MILP solvers in Multi-Agent Task Allocation and Scheduling Problems. Experimental results demonstrate that our method reduces optimization time and variance compared to traditional techniques while maintaining solution quality and feasibility.

\end{abstract}
\vspace{-3mm}
\section{INTRODUCTION}

Mixed Integer Linear Programs (MILPs) serve as a fundamental framework for combinatorial optimization problems, facilitating solutions across a wide range of planning and scheduling tasks in logistics~\citep{song_persistent_2018},  construction~\citep{huang2011optimization} and manufacturing~\citep{altundas_learning_2022}. These problems often involve making time-sensitive, resource-constrained decisions about what actions to take, when to take them, and how to coordinate them—challenges central to planning and scheduling problems in a wide range of industrial sectors~\citep{verderame2010planning}. As MILP aims to solve NP-hard problems such as Task Allocation and Scheduling~\citep{gombolay_fast_2018, wang_learning_2020}, there are significant challenges in terms of computation time, particularly for large-scale or time-sensitive applications~\citep{natarajan_human-robot_2023}. Traditional solution techniques used in the MILP Solvers, such as Branch-and-Bound (B\&B) and constraint generation, require computational resources to converge into an optimal solution~\citep{jimenez-cordero_warm-starting_2022}. To enable the practical deployment of intelligent robotic systems in construction environments, reducing solver latency becomes crucial.

Warm-starting has emerged as a promising strategy to accelerate MILP solvers by providing high-quality initial solutions, reducing the number of iterations needed for convergence~\citep{li2024faster}. In particular, BC—a supervised learning approach—has emerged as a promising method for learning policies replicating expert demonstrations in optimization tasks. RL also has been adopted to fine-tune BC's performance further. Our paper explores the use of BC and RL to warm-start MILPs for planning and scheduling applications in complex, real-world construction environments—reducing computational overhead while ensuring solution quality and operational reliability.

\section{Related Works}

\subsection{Classical Methods for Solving MILPs}

Warm-starting has long been used to accelerate MILP solvers by reusing information from previous instances. The general approach is to start each B\&B from the previous solver run's final B\&B leaves~\cite{gamrath2015reoptimization}. Ralphs et al. \citep{ralphs_duality_2006} proposed another method that utilizes previously computed B\&B trees and dual-derived information to efficiently re-solve new problem instances.

\subsection{Machine Learning for MILPs}

Machine learning has increasingly been integrated into MILP solving, both by enhancing solver internals and by generating useful external guidance. On the solver side, approaches include using generative models to learn branching policies that mimic strong branching decisions~\citep{jimenez-cordero_warm-starting_2022} and learning more effective primal heuristics~\citep{shen2021learning}. In addition, by leveraging partial or sub-optimal solutions generated by learning-based methods, previous work has shown to achieve state-of-the-art performance~\citep{huang_contrastive_2024}. BC, for instance, treats optimization trajectories as supervised learning datasets of state-action pairs, enabling models to imitate expert strategies~\citep{wang_learning_2020}. However, BC often struggles to generalize to unseen instances due to distribution shift. To mitigate this, online fine-tuning BC policies with RL has proven effective~\citep{zhu2018reinforcement, li2024faster}, allowing models to adapt based on solver feedback and improve performance in unfamiliar environments.

\subsection{Warm Starting in MILPs}

Recent works have proposed learning branching strategies~\citep{jimenez-cordero_warm-starting_2022} within MILP solvers. While promising, such an approach often requires tight integration with solver internals and needs to run model training multiple times, which is computationally heavy. Other works optimize via initializing solvers with expert-like solutions~\citep{song_persistent_2018, nair_solving_2021}, but it is limited in scalability guarantees~\citep{natarajan_human-robot_2023}. GNN has shown the ability to provide sub-optimal solutions at different scales for NP-Hard Problems such as task allocation and scheduling~\citep{wang_learning_2020, altundas_learning_2022}. In this paper, we will build on these advances and investigate the use of the BC+RL fine-tuning framework to train GNNs to warm-start MILP, which retains compatibility with off-the-shelf solvers and supports large-scale, temporally constrained multi-agent scheduling.

\section{Methodology}
\subsection{Multi-Agent Task Domain}

\begin{figure}
    \centering
    \includegraphics[width=0.9\linewidth]{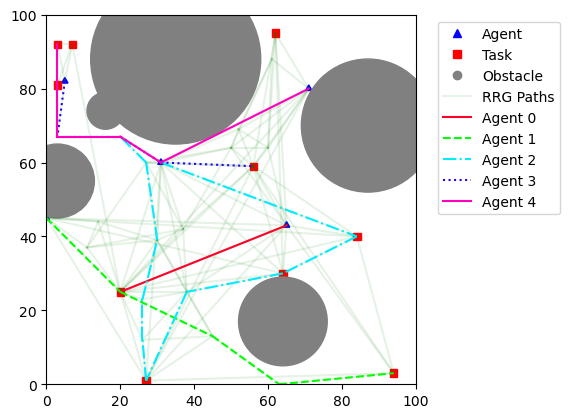}
    \vspace{-5mm}
    \caption{Multi-Agent Task Allocation and Motion Planning Environment showing the Agents, Tasks, Obstacles, and pre-computed Motion Planning through Multi-Agent Rapidly-exploring Random Graphs, with different agents moving along different paths to complete the assigned tasks.}
    \label{fig:environment with motion plan}
    \vspace{-6mm}
\end{figure}

We develop a simulation environment tailored to construction-inspired multi-agent task allocation and motion planning scenarios, an example shown in Figure~\ref{fig:environment with motion plan}, which tackles the challenge of optimizing long-term sequential decision-making in a continuous domain with obstacles. Our environment incorporates agents with heterogeneous velocities and task makespan to represent the heterogeneity present in real-world scenarios where agents possess varying capabilities.

\subsection{Environment Design}
The continuous environment is structured to simulate realistic construction scenarios, including the presence of obstacles that agents must navigate and the order constraints between tasks and time windows. The obstacles and constraints lead to challenges in both task allocation and motion planning, as agents must consider not only the optimal task sequence with respect to constraints but also the feasibility and efficiency of their paths.

\begin{itemize}
    \item \textbf{Order Constraint:} Some tasks must be executed in a predefined order due to dependencies. Violating these constraints would result in infeasibility of the task execution.
    \item \textbf{Time Window:} Each task is associated with a specific time window within which it must be completed. Tasks that fall outside their designated time windows are deemed infeasible.
\end{itemize}

To estimate travel times between task locations, we precompute collision-free paths using Multi-Agent Rapidly-exploring Random Graphs (MA-RRG), allowing agents to navigate around static obstacles efficiently~\cite{kala2013rapidly}.



\subsection{MILP Approach for Task Scheduling and Assignment}
\label{appendix:MILP_Solver}

To generate a task schedule that balances agent capabilities and travel times, we adopt a MILP formulation that jointly solves multi-agent task assignment and scheduling.
The constraints used in our MILP formulation for Multi-Agent Task Allocation and Scheduling are presented in Equation~\ref{eq:MILP}, with constraints (C1) through (C10). The MILP constraint is that each task is assigned to exactly one agent and completed once. Two tasks assigned to the same agent can only be completed in a single order, (C2) and (C3), through sequencing constraints. Agents cannot move from one task to the next without completing the previous tasks and have to travel through a predefined path for a pre-calculated duration. The order of tasks obeys the wait time constraints and time windows, constraints (C6), (C7), (C8), and (C10). Finally, tasks take a duration determined by the heterogeneous task durations for different agents (C9).

The objective is to generate an optimized schedule by minimizing a cost function. Our methodology builds upon~\citep{gombolay_fast_2018}, incorporating heterogeneous travel times based on varying agent speeds and precomputed distances between initial agent locations and task sites.
\vspace{-1mm}
\begin{equation}
\label{eq:MILP}
    \begin{aligned}
    \min \quad & f(\mathbf{A}, \mathbf{S}^1, \dots, \mathbf{S}^{N_A}) \\
    \text{s.t.} \quad 
    & \sum_{i\in\mathcal{A}} \mathbf{A}_{ij} = 1 \quad \forall j\in\mathcal{T} && \makebox[0pt][r]{(C1)} \\
    & \mathbf{S}^i_{jk} + \mathbf{S}^i_{kj} \leq  \mathbf{A}_{ij} \quad \forall j, k \in\mathcal{T}, i\in\mathcal{A} && \makebox[0pt][r]{(C2)} \\
    & \mathbf{S}^i_{jk} + \mathbf{S}^i_{kj} \leq \mathbf{A}_{ik}\quad \forall j, k \in\mathcal{T}, i\in\mathcal{A} && \makebox[0pt][r]{(C3)} \\
    & t^A_k \geq -M \left( 3 - \left( \mathbf{A}_{ij} + \mathbf{A}_{ik} + \mathbf{S}^i_{jk} \right) \right) + t^F_j + t^T_{ijk} \\
    & \quad \quad \quad \quad \forall j, k \in\mathcal{T}, i\in\mathcal{A} && \makebox[0pt][r]{(C4)} \\ 
    & t^A_k \geq -M(1-\mathbf{A}_{ik}) + t^T_{ik} \quad \forall k \in\mathcal{T}, i\in\mathcal{A} && \makebox[0pt][r]{(C5)} \\
    & t^S_k \geq t^A_k, \forall k \in\mathcal{T} && \makebox[0pt][r]{(C6)} \\
    & t^S_k \geq M(\mathbf{O}_{jk}-1) + t^F_j + W_{jk} \quad \forall j, k \in\mathcal{T} && \makebox[0pt][r]{(C7)} \\
    & t^S_k \geq s_k \quad, \forall k \in\mathcal{T} && \makebox[0pt][r]{(C8)} \\
    & t^F_k \geq t^S_k + \mathbf{A}_{ik} t^E_{ik} \quad \forall k \in\mathcal{T}, i\in\mathcal{A} && \makebox[0pt][r]{(C9)} \\ 
    & t^F_k \leq e_k \quad \forall k \in\mathcal{T} && \makebox[0pt][r]{(C10)}
    \end{aligned}
\end{equation}

where $M$ is a sufficiently large positive constant used to enforce conditional constraints. 

\vspace{-2mm}
\subsection{Learning-Based Methods}

\subsubsection{Behavior Cloning}
BC is employed to accelerate the optimization process by leveraging expert-generated data to warm-start the MILP solver. Expert MILP solvers generate high-quality solutions for 200 environment instances focused on scenarios with 10 agents and 20 tasks. These outputs are preprocessed to match the input format required by the GNN model. The GNN is then trained to imitate the MILP solver by minimizing the discrepancy between its predictions and expert actions. This trained GNN is evaluated on the same scaled test environments to assess generalization. By providing high-quality initializations, the BC approach significantly speeds up the MILP solver and enables broader deployment across similar task allocation problems.



\subsubsection{Fine-tuning via Reinforcement Learning}

RL further refines task allocation by training the GNN through interaction with the environment and MILP solver. The model generates task assignments, which are refined by the solver and used to update the GNN via reward-driven learning. Rewards combine schedule quality and time required to optimize. The schedule quality score is calculated at the final time step ($t=|T|$) determined by number of Tasks, that penalizes both makespan and infeasible assignments as defined in Equations~\eqref{eq:r1}, where $t_{ms}$ is the total makespan, $t_{ddl}$ is the total deadline, and $\mathbb{1}_{\text{feasible}^i}$ means task assignment for task $i$ can be done within MILP's constraints. The optimization time required to optimize is defined in Equations~\eqref{eq:r2}. The design of this reward structure is still being explored to maximize effectiveness.

\vspace{-4mm}
\begin{equation}
R_{score} = \frac{\sum_{i=1}^{|T|} \mathbb{1}_{\text{feasible}^i} + \left(1.0 - \frac{t_{ms}}{t_{ddl}}\right)}{T+1}
\label{eq:r1}
\vspace{-2mm}
\end{equation}
\begin{equation}
R_{time} = - t_{optimization}
\label{eq:r2}
\vspace{-4mm}
\end{equation}

\begin{figure}
    \centering
    \includegraphics[width=1.0\linewidth]{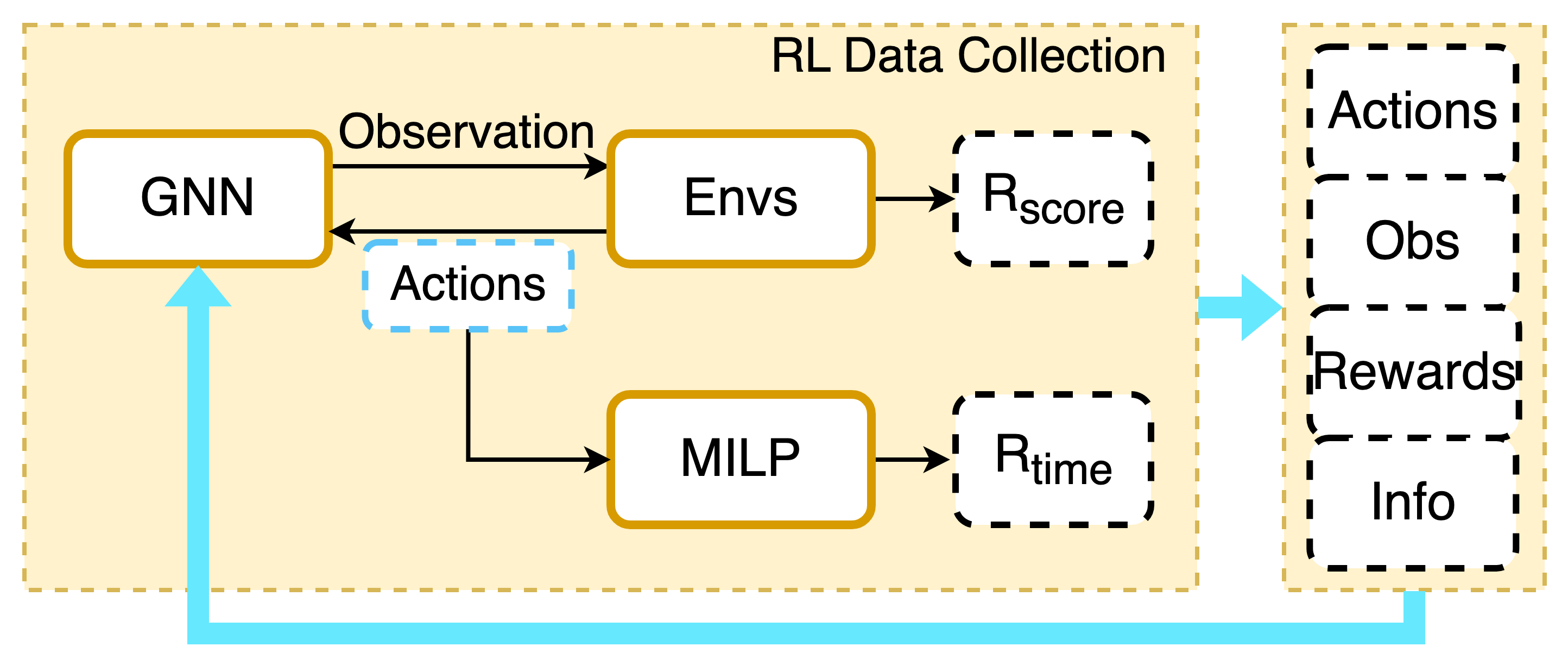}
    \vspace{-8mm}
    \caption{Reinforcement Learning framework showing the interaction between the GNN model, environments, and the warm-started MILP solver.}
    \label{fig:rl framework}
    \vspace{-5mm}
\end{figure}

\subsection{Warm-starting Optimizer with Schedules}
The MILP solver is designed to initialize each time it is called, consistently reading from the same set of parameter settings. This consistency ensures that the MILP solver’s performance remains stable across different optimization runs, providing a reliable baseline for evaluation.

To test the performance of different methods, partially optimized task allocation schedules obtained from various methods are passed into the MILP solver as initial solutions. The optimization time required to achieve the optimal solution is recorded as a key performance metric. By comparing the performance of different warm-starting methods, the effectiveness of each approach can be evaluated. The consistency of the MILP solver's initialization process ensures that any observed improvements can be attributed to the quality of the provided initial schedules rather than stochastic variations in the solver's behavior.

\section{Experimental Results}
To evaluate the proposed approach, we conducted experiments on our Multi-Agent Task Allocation and Motion Planning environment described in Section III. Our focus is to compare the efficiency of various methods, including \textbf{Baseline (Exact Solvers without warm-start)}, \textbf{EDF (Earliest Deadline First)}, \textbf{Constraint-Aware EDF}, \textbf{Behavior Cloning (BC Only)}, and \textbf{Behavior Cloning with Reinforcement Learning Fine-Tuning (BC RL)}.

EDF is a heuristic that assigns tasks to robots greedily based on the earliest deadline, aiming to minimize task completion time without considering task dependencies~\citep{kargahi2006method}. Constraint-Aware EDF extends this by accounting for task prerequisites, only scheduling tasks whose dependencies have been satisfied~\citep{kartal2016monte}.

\subsection{Experimental Setup and Metrics}


We evaluate five methods (Baseline, EDF, Constraint-Aware EDF, BC Only, and BC RL) on 100 randomly generated instances with 10 agents and 20 tasks. Each instance runs 10 times. We report the average and standard deviation of optimization time and quality score. Time measures solver speed; quality scores combine feasibility and makespan relative to deadlines, higher scores indicate more feasible and compact schedules, with values ranging from 1 to 21.









\vspace{-2mm}
\subsection{Results Analysis}

\begin{figure}
    \centering
    \includegraphics[width=1.0\linewidth]{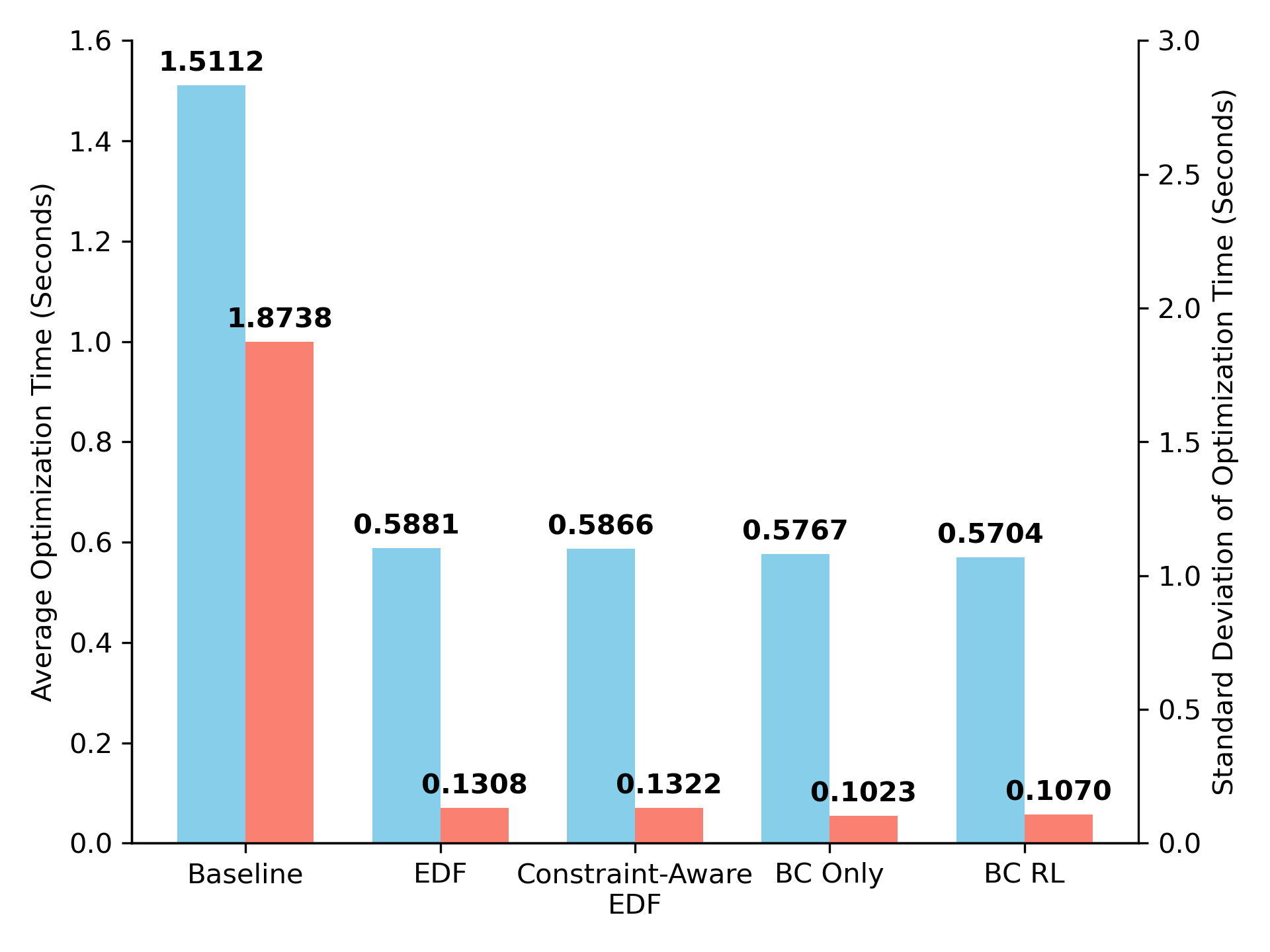}
    \vspace{-10mm}
    \caption{Average and standard deviation of optimization time in 10 agents 20 tasks domain across methods. All warm-start methods reduce time; BC RL achieves the lowest average time with stable performance.}
    \label{fig:10r20t avg and std time}
    \vspace{-3mm}
\end{figure}

\begin{figure}
    \centering
    \includegraphics[width=1.0\linewidth]{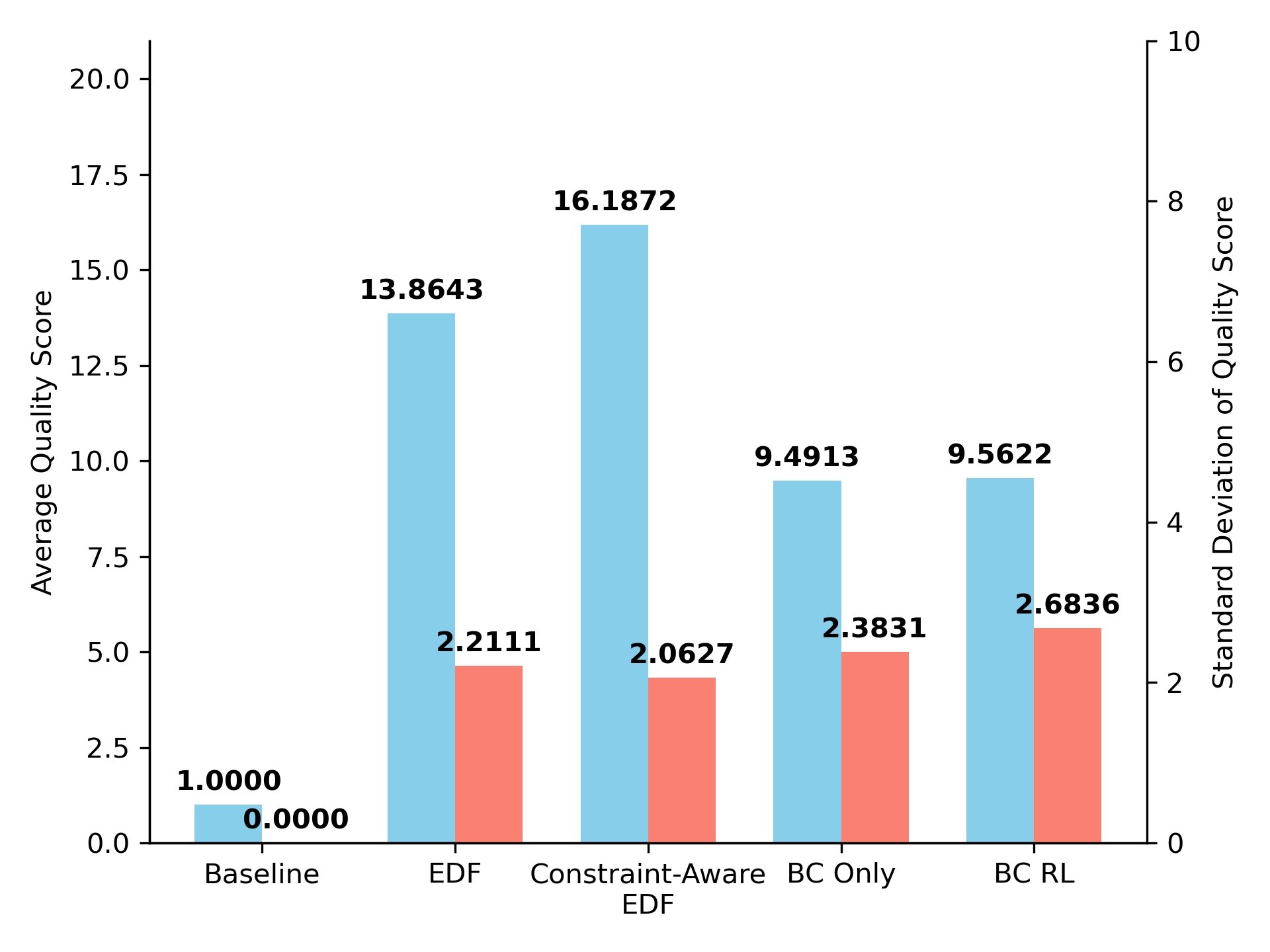}
    \vspace{-10mm}
    \caption{Average and standard deviation of quality score in 10 agents 20 tasks domain across methods. BC RL achieves competitive quality with low variance, balancing performance and stability.}
    \label{fig:10r20t avg and std score}
    \vspace{-6mm}
\end{figure}

The results in Figure~\ref{fig:10r20t avg and std time} show that all learning-based methods significantly outperform the baseline in terms of optimization time. BC achieves better performance over EDF and Constraint-Aware EDF, with the additional fine-tuning via RL providing slight improvements. The standard deviation values indicate low variance, demonstrating the robustness of the proposed methods. The current measure of optimization time includes searching time and validation time. The lack of significant improvement in the optimization time for BC and BC+RL may be due to this. Searching time will be analyzed separately in the future.

In Figure~\ref{fig:10r20t avg and std score} Constraint-Aware EDF achieved the highest average quality score, followed by EDF, indicating strong schedule feasibility and compactness. The learning-based methods, BC with RL and BC Only, outperformed the Baseline but were slightly less effective than EDF-based heuristics. This may be due to the EDF-based heuristics explicitly enforcing temporal and dependency constraints during task assignment. In terms of stability, Constraint-Aware EDF had the lowest standard deviation, while BC RL showed the most variability, suggesting occasional performance inconsistency.

\vspace{-1mm}
\section{Discussion and Conclusion}
\vspace{-1mm}
Our results show that BC effectively warm-starts MILP solvers, significantly reducing optimization time and outperforming the baseline in quality. RL fine-tuning provides modest gains on unseen instances but lacks consistency, indicating room for improvement in training and reward design. While learning-based methods perform well, they still fall short of heuristic methods like Constraint-Aware EDF in peak quality and stability. BC learns to provide solutions by minimizing cross-entropy between the optimal solution and the learned model, which may allow it to learn to output solutions that are closer to the optimal solution despite having a lower performance, making search take shorter times compared to EDF-based heuristics~\cite{goecks2019integrating}. RL optimizes BC further via online fine-tuning to take less time.

Future work will focus on refining RL training and reward design, as well as testing scalability to larger instances (e.g., 20 agents, 100 tasks) to assess the robustness of our approach. Fidelity will also be used to investigate how close the policies we learn are to the optimal ones. It is also worth exploring ways of combining the study of branching policy and warm-start to get both good branching strategies and good bounding policy.
\vspace{-6mm}

\bibliographystyle{IEEEtran}
\bibliography{references}

\begin{thebibliography}{10}
\providecommand{\url}[1]{#1}
\csname url@samestyle\endcsname
\providecommand{\newblock}{\relax}
\providecommand{\bibinfo}[2]{#2}
\providecommand{\BIBentrySTDinterwordspacing}{\spaceskip=0pt\relax}
\providecommand{\BIBentryALTinterwordstretchfactor}{4}
\providecommand{\BIBentryALTinterwordspacing}{\spaceskip=\fontdimen2\font plus
\BIBentryALTinterwordstretchfactor\fontdimen3\font minus \fontdimen4\font\relax}
\providecommand{\BIBforeignlanguage}[2]{{%
\expandafter\ifx\csname l@#1\endcsname\relax
\typeout{** WARNING: IEEEtran.bst: No hyphenation pattern has been}%
\typeout{** loaded for the language `#1'. Using the pattern for}%
\typeout{** the default language instead.}%
\else
\language=\csname l@#1\endcsname
\fi
#2}}
\providecommand{\BIBdecl}{\relax}
\BIBdecl

\bibitem{song_persistent_2018}
B.~D. Song, K.~Park, and J.~Kim, ``Persistent {UAV} delivery logistics: {MILP} formulation and efficient heuristic,'' \emph{Computers \& Industrial Engineering}, vol. 120, pp. 418--428, 2018, publisher: Elsevier.

\bibitem{huang2011optimization}
C.~Huang, C.~K. Wong, and C.~M. Tam, ``Optimization of tower crane and material supply locations in a high-rise building site by mixed-integer linear programming,'' \emph{Automation in Construction}, vol.~20, no.~5, pp. 571--580, 2011.

\bibitem{altundas_learning_2022}
\BIBentryALTinterwordspacing
B.~Altundas, Z.~Wang, J.~Bishop, and M.~Gombolay, ``Learning {Coordination} {Policies} over {Heterogeneous} {Graphs} for {Human}-{Robot} {Teams} via {Recurrent} {Neural} {Schedule} {Propagation},'' in \emph{2022 {IEEE}/{RSJ} {International} {Conference} on {Intelligent} {Robots} and {Systems} ({IROS})}.\hskip 1em plus 0.5em minus 0.4em\relax Kyoto, Japan: IEEE, Oct. 2022, pp. 11\,679--11\,686. [Online]. Available: \url{https://ieeexplore.ieee.org/document/9981748/}
\BIBentrySTDinterwordspacing

\bibitem{verderame2010planning}
P.~M. Verderame, J.~A. Elia, J.~Li, and C.~A. Floudas, ``Planning and scheduling under uncertainty: a review across multiple sectors,'' \emph{Industrial \& engineering chemistry research}, vol.~49, no.~9, pp. 3993--4017, 2010.

\bibitem{gombolay_fast_2018}
M.~C. Gombolay, R.~J. Wilcox, and J.~A. Shah, ``Fast scheduling of robot teams performing tasks with temporospatial constraints,'' \emph{IEEE Transactions on Robotics}, vol.~34, no.~1, pp. 220--239, 2018, publisher: IEEE.

\bibitem{wang_learning_2020}
Z.~Wang and M.~Gombolay, ``Learning scheduling policies for multi-robot coordination with graph attention networks,'' \emph{IEEE Robotics and Automation Letters}, vol.~5, no.~3, pp. 4509--4516, 2020, publisher: IEEE.

\bibitem{natarajan_human-robot_2023}
\BIBentryALTinterwordspacing
M.~Natarajan, E.~Seraj, B.~Altundas, R.~Paleja, S.~Ye, L.~Chen, R.~Jensen, K.~C. Chang, and M.~Gombolay, ``\BIBforeignlanguage{en}{Human-{Robot} {Teaming}: {Grand} {Challenges}},'' \emph{\BIBforeignlanguage{en}{Current Robotics Reports}}, vol.~4, no.~3, pp. 81--100, Aug. 2023. [Online]. Available: \url{https://link.springer.com/10.1007/s43154-023-00103-1}
\BIBentrySTDinterwordspacing

\bibitem{jimenez-cordero_warm-starting_2022}
A.~Jiménez-Cordero, J.~M. Morales, and S.~Pineda, ``Warm-starting constraint generation for mixed-integer optimization: {A} machine learning approach,'' \emph{Knowledge-Based Systems}, vol. 253, p. 109570, 2022, publisher: Elsevier.

\bibitem{li2024faster}
Z.~Li, L.~Chen, R.~Paleja, S.~Nageshrao, and M.~Gombolay, ``Faster model predictive control via self-supervised initialization learning,'' \emph{arXiv preprint arXiv:2408.03394}, 2024.

\bibitem{gamrath2015reoptimization}
G.~Gamrath, B.~Hiller, and J.~Witzig, ``Reoptimization techniques for mip solvers,'' in \emph{International Symposium on Experimental Algorithms}.\hskip 1em plus 0.5em minus 0.4em\relax Springer, 2015, pp. 181--192.

\bibitem{ralphs_duality_2006}
T.~Ralphs and M.~Güzelsoy, ``Duality and warm starting in integer programming,'' 2006.

\bibitem{shen2021learning}
Y.~Shen, Y.~Sun, A.~Eberhard, and X.~Li, ``Learning primal heuristics for mixed integer programs,'' in \emph{2021 international joint conference on neural networks (ijcnn)}.\hskip 1em plus 0.5em minus 0.4em\relax IEEE, 2021, pp. 1--8.

\bibitem{huang_contrastive_2024}
T.~Huang, A.~M. Ferber, A.~Zharmagambetov, Y.~Tian, and B.~Dilkina, ``Contrastive predict-and-search for mixed integer linear programs,'' 2024.

\bibitem{zhu2018reinforcement}
Y.~Zhu, Z.~Wang, J.~Merel, A.~Rusu, T.~Erez, S.~Cabi, S.~Tunyasuvunakool, J.~Kram{\'a}r, R.~Hadsell, N.~de~Freitas \emph{et~al.}, ``Reinforcement and imitation learning for diverse visuomotor skills,'' \emph{arXiv preprint arXiv:1802.09564}, 2018.

\bibitem{nair_solving_2021}
\BIBentryALTinterwordspacing
V.~Nair, S.~Bartunov, F.~Gimeno, I.~v. Glehn, P.~Lichocki, I.~Lobov, B.~O'Donoghue, N.~Sonnerat, C.~Tjandraatmadja, P.~Wang, R.~Addanki, T.~Hapuarachchi, T.~Keck, J.~Keeling, P.~Kohli, I.~Ktena, Y.~Li, O.~Vinyals, and Y.~Zwols, ``Solving {Mixed} {Integer} {Programs} {Using} {Neural} {Networks},'' Jul. 2021, arXiv:2012.13349 [math]. [Online]. Available: \url{http://arxiv.org/abs/2012.13349}
\BIBentrySTDinterwordspacing

\bibitem{kala2013rapidly}
R.~Kala, ``Rapidly exploring random graphs: motion planning of multiple mobile robots,'' \emph{Advanced Robotics}, vol.~27, no.~14, pp. 1113--1122, 2013.

\bibitem{kargahi2006method}
M.~Kargahi and A.~Movaghar, ``A method for performance analysis of earliest-deadline-first scheduling policy,'' \emph{The Journal of Supercomputing}, vol.~37, pp. 197--222, 2006.

\bibitem{kartal2016monte}
B.~Kartal, E.~Nunes, J.~Godoy, and M.~Gini, ``Monte carlo tree search for multi-robot task allocation,'' in \emph{Proceedings of the AAAI Conference on Artificial Intelligence}, vol.~30, no.~1, 2016.

\bibitem{goecks2019integrating}
V.~G. Goecks, G.~M. Gremillion, V.~J. Lawhern, J.~Valasek, and N.~R. Waytowich, ``Integrating behavior cloning and reinforcement learning for improved performance in dense and sparse reward environments,'' \emph{arXiv preprint arXiv:1910.04281}, 2019.

\end{thebibliography}





\end{document}